\documentclass[letterpaper, 10 pt, conference]{ieeeconf}  

\IEEEoverridecommandlockouts  

\usepackage{hyperref}
\usepackage{amsmath}
\usepackage{graphicx}
\usepackage{makecell}
\usepackage{subcaption}
\usepackage{multirow} 

\graphicspath{{Images/}}

\title{\LARGE \bf
Collision-Free Robot Navigation in Crowded Environments using Learning based Convex Model Predictive Control
}

\author{Zhuanglei Wen, Mingze Dong, and Xiai Chen
\thanks{*Corresponding author: Mingze Dong}
\thanks{All authors are with the College of Mechanical and Electrical Engineering, China Jiliang University, Hangzhou 310018, Zhejiang, China
        {\tt\small 1801400116@cjlu.edu.cn}}%
}

\begin{document}

\maketitle
\thispagestyle{empty}
\pagestyle{empty}

\begin{abstract}
Navigating robots safely and efficiently in crowded and complex environments remains a significant challenge. However, due to the dynamic and intricate nature of these settings, planning efficient and collision-free paths for robots to track is particularly difficult. In this paper, we uniquely bridge the robot’s perception, decision-making and control processes by utilizing the convex obstacle-free region computed from 2D LiDAR data. The overall pipeline is threefold: (1) We proposes a robot navigation framework that utilizes deep reinforcement learning (DRL), conceptualizing the observation as the convex obstacle-free region, a departure from general reliance on raw sensor inputs. (2) We design the action space, derived from the intersection of the robot’s kinematic limits and the convex region, to enable efficient sampling of inherently collision-free reference points. These actions assists in guiding the robot to move towards the goal and interact with other obstacles during navigation. (3) We employ model predictive control (MPC) to track the trajectory formed by the reference points while satisfying constraints imposed by the convex obstacle-free region and the robot’s kinodynamic limits. The effectiveness of proposed improvements has been validated through two sets of ablation studies and a comparative experiment against the Timed Elastic Band (TEB), demonstrating improved navigation performance in crowded and complex environments. 
\end{abstract}

\section{INTRODUCTION}

Robot navigation in crowded environments is still a challenge that has drawn significant attention from the global research community \cite{obstacle_crowd}. This interest is heightened by the integration of advanced learning methodologies \cite{autonomous_navigation_challenge_1}, \cite{autonomous_navigation_challenge_2}. A reliable and effective autonomous navigation strategy is required due to the unpredictable dynamics. Recent advances in machine learning and DRL have facilitated the investigation of neural networks for navigation in these challenging settings \cite{DRL_1}, \cite{DRL_2}, \cite{DRL_3}.

Current DRL-based end to end navigation strategies typically define the expected speed, acceleration, and other control variables in different dimensions as actions, within continuous action spaces \cite{continuous_1}, \cite{continuous_2}. Alternatively, these control variables are integrated into predefined actions (e.g., moving forward, braking) to form discrete action spaces \cite{discrete_1}, \cite{discrete_2}. Nevertheless, neither continuous nor discrete actions directly depict the robot's expected trajectory. Akmandor et al. \cite{Akmandor} proposed mapping control variables to a set of motion primitives, enabling the agent to select the optimal trajectory to track. Nevertheless, these methods still cannot guarantee the interpretability and safety of the robot's motion, losing explicit mechanism for the integration of optimization/decision and control, thereby leading to uninterpretable autonomy and impractical implementation. To address this challenge, researchers, including Gao et al. \cite{Gao}, have introduced innovative approaches. One effective approach is to separate the control mechanism from the neural network and employ control methods that are feasible in real-world settings.
\begin{figure}[htbp]
    \centering
    \includegraphics[width=8cm]{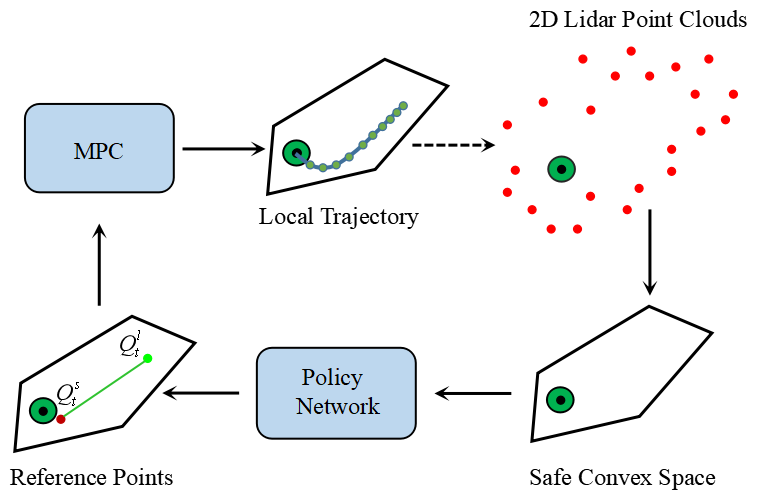}
    \caption{Proposed Navigation Architecture: The convex obstacle-free region is obtained from 2D LiDAR point cloud data. The policy network selects reference points (\(Q_{t}^{s}\) and \(Q_{t}^{l}\)) from this convex region based on consecutive frames of observations. The MPC is then employed with online optimization to compute optimal local trajectory. The trajectory follows the reference points closely while satisfying both kinodynamic constraints and convex obstacle-free region requirements. This iterative process continues until the robot reaches its goal. }
    \label{fig:Proposed Navigation Architecture}
\end{figure}

Inspired by these advancements, our study employs MPC, valued for its ability to enforce hard constraints, ensuring control commands stay within predefined limits \cite{MPC_constraint}. Consequently, we define the action as a sequence of reference points for MPC's reference trajectory. Furthermore, we utilize the convex obstacle-free region derived from LiDAR point clouds as the constraint space, as depicted in Fig. \ref{fig:Proposed Navigation Architecture}. This strategy confines the MPC-optimized trajectory strictly within the convex region, thereby ensuring navigation that is both kinematically feasible and safe.

The principal contributions of this study are as follows:

\begin{itemize}
\item
  We design the action and action space based on the convex obstacle-free region derived from LiDAR data. To facilitate both short-term dynamic obstacle avoidance and long-term navigation, we characterize the action as a reference trajectory composed of reference points at specific time intervals. The action space is defined as the intersection of the convex region with the robot's motion limits, from which we sample collision-free reference points efficiently. Subsequently, MPC is employed to track the trajectory, incorporating the convex region constraint within the MPC formulation to ensure the generation of safe and reliable control commands. Furthermore, we develop customized state space and reward function based on the convex region, reference points, and MPC-optimized trajectory.
  %
  
\item
 We implement a seven-stage curriculum training strategy and evaluate it through detailed experiments. The experiments include ablation studies to analyze the impact of varying action spaces and reward functions on navigation performance. Furthermore, a comparative evaluation with TEB method highlighted the superior performance of our method.
\item
  We have made our work open source by publishing both the implementation and benchmark data\footnote{ \url{https://github.com/sunnyhello369/flat_ped_sim.git}}.
\end{itemize}

\section{RELATED WORK}

Previous robot navigation approaches often consider velocity, force, potential, and other factors as key concepts \cite{velocity-based_force-based_potential}. However, they face challenges due to their reliance on simplistic assumptions about pedestrian dynamics. The Social Force Model (SFM) \cite{SFM}, along with methods like Reciprocal Velocity Obstacles (RVO) \cite{RVO} and Optimal Reciprocal Collision Avoidance (ORCA) \cite{ORCA}, provide foundational frameworks for predicting pedestrian movements, yet they may not fully capture the unpredictability of real-world environments \cite{old_methods_flaw}. Chen et al. \cite{iMPC} proposed an interactive Model Predictive Control (iMPC) framework that utilizes the iORCA model for enhanced prediction of pedestrian movements, thereby improving robot navigation in crowded environments.

Learning based methods are also applied to navigation in crowded environments. Yao et al. \cite{Yao} developed an end-to-end DRL framework that uses sensor data and pedestrian maps to distinguish between obstacles and pedestrians to enhance dynamic obstacle avoidance capabilities. Hu et al. \cite{hu_drl_transformer} presented a planning-oriented framework that utilizes Transformer-based models \cite{transformer} for perception, prediction, and planning tasks. This framework overcomes the limitations of modular designs and multi-task learning, achieving end-to-end navigation. 

In parallel, hybrid methods have also seen novel applications. Gao et al. \cite{Gao} developed a method, encoding vehicle and predicted trajectories into binary images as the state. Action is defined as vehicle's future positions in polar coordinate. They applied convex optimization to these discrete reference points, transforming them into reachable trajectory to track. This method decouples control from other processes, with DRL focuses on suggesting reference positions and convex optimization ensuring the trajectory's feasibility for vehicle execution.

Nikdel et al. \cite{lbgp_following_in_front} introduced a hybrid method that combines DRL with classical trajectory planning, where DRL estimates human trajectories and suggests short-term goals. These goals are then used by TEB to navigate the robot in front of humans. Linh et al. \cite{Linh} had explored various planning methods, including MPC, TEB, and DRL-based end-to-end navigation strategies. They designed policy network as a strategy selector, choosing the appropriate strategy based on static and dynamic environmental information. Brito et al. \cite{Go_MPC} proposed a DRL-based decision-planning method that recommends target positions for the MPC planner. Facilitate navigation by considering interactions with other agents. Compared to \cite{Go_MPC}, our method does not require states of obstacles to be observed, only requires raw 2D LiDAR point cloud data.

Our navigation strategy is similar to that of Gao et al. \cite{Gao}, where the action is defined as a sequence of future positions for the robot. However, to achieve better navigation performance and smoother motion trajectories in dynamic crowded environments, our method utilizes the state observation based on the convex obstacle-free region. Furthermore, we integrate the convex region into the architecture of the action-state space, the reward function, and our MPC framework.

\section{METHOD}

\subsection{Problem Definition} 
We model our problem using a Partially Observable Markov Decision Process (POMDP), which is well-suited for environments with inherent uncertainties and dynamic conditions. The POMDP model \(M = (S, A, T, R, \Omega, O, \gamma)\) consists of state space \(S\), action space \(A\), state transition function \(T\), reward function \(R\), finite set of observations \(\Omega\), observation function \(O\), and discount factor \(\gamma \in [0, 1)\). At time \(t\), the action \(a_t\), consists of the short-term and long-term reference points, \(Q_{t}^{s}\) and \(Q_{t}^{l}\), is formulated in Section~\ref{subsec:Action Space} and the reward \(r_t\) detailed in Section~\ref{subsec:Reward Function}. The state \(s_t\) is defined in Section \ref{subsec:Observation Space}.

\subsection{Agent Dynamics} \label{subsec:Agent Dynamics}
To facilitate computation in planning and control, the omni-directional mobile robot is simplified to a point mass. This simplification converts the robot's kinematics into a 2D third-order integral model. The state of the model denoted by \(x=\left[ p_x,p_y,v_x,v_y,a_x,a_y \right] ^T\), represents the position, velocity, and acceleration. On the other hand, the control input, \(u = \left[ j_x, j_y \right] ^T\), denotes the jerk.

The state evolution of the system can be expressed as:
\begin{align}
\dot{p}_x &= v_x & \dot{p}_y &= v_y \notag \\
\dot{v}_x &= a_x & \dot{v}_y &= a_y \notag \\
\dot{a}_x &= j_x & \dot{a}_y &= j_y
\label{eq:Agent Dynamics}
\end{align}

\subsection{Action Space Design Based on Convex Region}\label{subsec:Action Space}
The convex obstacle-free region is generated by the algorithm proposed by Zhong et al. \cite{zhong_generate_convex}, which can efficiently create reliable convex spaces among obstacles of any shape from LiDAR point cloud data.

To achieve efficient dynamic obstacle avoidance, the agent's action is designed as a sequence of reference points at specific time intervals. In order to simplify the training process, this sequence is simplified to two points: one that represents the robot's reference position after a control cycle (\(t_c\)) and another that represents after an MPC prediction horizon (\(T\)). These points serve as the short-term and long-term reference points, respectively. The path from the robot's current position to these points forms the reference trajectory for MPC tracking.

\begin{figure}[htbp]
    \centering
    \includegraphics[width=10cm]{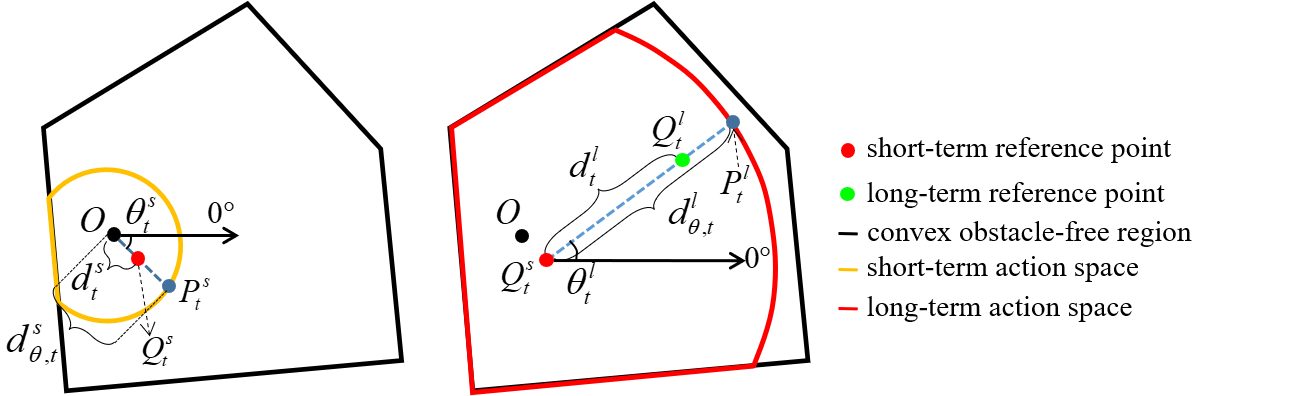}
    \caption{Schematic of Short-Term and Long-Term Reference Point Formulation}
    \label{fig:Short-Term and Long-Term Reference Point}
\end{figure}

When designing the action space, we first ensure that the sampling of reference points is confined within the convex region. Then, centering on the robot's current position \(O\) and considering its kinematic limits, we define two circular areas. These circles, intersected with the convex region, represent our short-term and long-term action spaces, as illustrated by Fig. \ref{fig:Short-Term and Long-Term Reference Point}. The policy network is designed to output two sets of two-dimensional data. These data, after sigmoid activation, yield \(\hat{a}_t=\left\{ \left( \alpha _{t}^{s},\beta _{t}^{s} \right) ,\left( \alpha _{t}^{l},\beta _{t}^{l} \right) |\alpha _t,\beta _t\in \left( 0,1 \right) \right\} \).

The short-term and long-term reference points coordinates \(Q_{t}^{s}\) and \(Q_{t}^{l}\), are calculated as Equation \ref{eq:short_term_and_long_term_reference_point}, with Fig. \ref{fig:Short-Term and Long-Term Reference Point} illustrating this process. A ray at angle \(\theta _{t}^{s}\) intersects with the short-term action space at \(P_{t}^{s}\). We determine \(P_{t}^{s}\) by calculating the polar angles of the convex's vertices and finding which edge the angle \(\theta _{t}^{s}\) falls within. This intersection, is solved by combining the ray and the edge equations, yields \(P_{t}^{s}\). The distance \(d_{\theta, t}^{s}\), from \(O\) to \(P_{t}^{s}\), is multiplied by the scaling factor \(\beta_{t}^{s}\) to yield the distance \(d_{t}^{s}\). Then we establish a new polar coordinate system centered at the short-term reference point \(Q_{t}^{s}\). The long-term reference angle \(\theta _{t}^{l}\), is obtained by adding the increment angle \(\alpha _{t}^{l}\cdot 2\pi\) to \(\theta_{t}^{s}\). Following the procedure used for the short-term point, we can determine the coordinates of \(Q_{t}^{l}\).
\begin{equation}
\begin{aligned}
\theta _{t}^{s}&=\alpha _{t}^{s}\cdot 2\pi \\
d_{t}^{s}&=\beta _{t}^{s}\cdot d_{\theta ,t}^{s} \\
Q_{t}^{s}&=\left( d_{t}^{s}\cdot \cos \left( \theta _{t}^{s} \right), d_{t}^{s}\cdot \sin \left( \theta _{t}^{s} \right) \right) \\
\theta _{t}^{l} &= \theta _{t}^{s}+\alpha _{t}^{l}\cdot 2\pi \\
d_{t}^{l} &= \beta _{t}^{l}\cdot d_{\theta ,t}^{l} \\
Q_{t}^{l} &= Q_{t}^{s}+\left( d_{t}^{l}\cdot \cos \left( \theta _{t}^{l} \right) ,d_{t}^{l}\cdot \sin \left( \theta _{t}^{l} \right) \right)
\end{aligned}
\label{eq:short_term_and_long_term_reference_point}
\end{equation}

\begin{figure}[htbp]
    \centering
    \includegraphics[width=8.5cm]{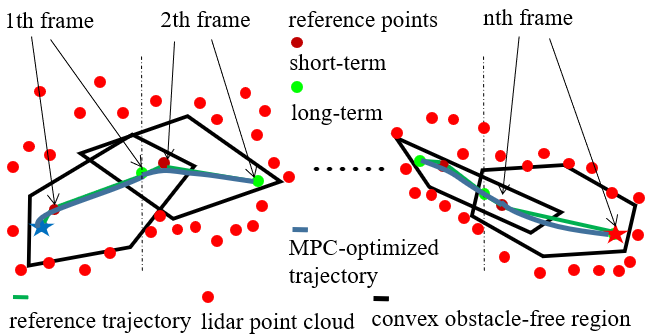}
    \caption{Schematic of Iterative Trajectory Optimization within the Convex Region}
    \label{fig:Iterative Trajectory Optimization}
\end{figure}

To navigate safely in unknown environment, our method utilizes real-time LiDAR data to construct the convex obstacle-free region for local navigation. By keeping the reference trajectory within the latest convex region, we can greatly reduce the risk of colliding with obstacles. The reference trajectory, connecting both long-term and short-term reference points is depicted by the green solid line in Fig. \ref{fig:Iterative Trajectory Optimization}. Subsequently, the MPC framework solves the local navigation problem by calculating optimal control inputs so that the robot can continuously follow the reference trajectory. This process is iterated, ensuring the robot can follow the updated trajectory and reach the target without colliding with obstacles.

\subsection{Model Predictive Control
Formulation}\label{subsec:MPC}
Let \(t_c\) represent the control period. If the MPC forecasts the system's state over N future control periods, the total prediction duration is \(T=N\cdot t_c\). The initial state \(x_{init}=\left[ p_{x}^{0},p_{y}^{0},v_{x}^{0},v_{y}^{0},a_{x}^{0},a_{y}^{0} \right] ^T\) reflects the observed state at the start of the period.

After discretizing the integral model that governs robotic dynamics (as detailed in Section \ref{subsec:Agent Dynamics}), we derive a sequential relationship expressed as \(x_i = Fx_{i-1} + Gu_{i-1}\) for \(i = \{1, \ldots, N\}\). In this context, \(F\) represents the matrix for discrete-time state transitions, and \(G\) refers to the matrix for discrete-time control inputs.

Using the method mentioned in Section~\ref{subsec:Action Space}, we get the convex region from point cloud data, with its vertices \(P_{j}^{c}, j=\left\{ 1,...,rnum_v \right\} \) arranged in clockwise direction. \(rnum_v\) is a hyperparameter that describes the fixed number of vertices and will be introduced at Section~\ref{subsec:Observation Space}. This region constrains the MPC-optimized points to enhance safety by reducing the risk of collisions with obstacles. This constraint is efficiently met by determining whether the trajectory point \(Q_i\) is inside the convex region through the vector cross product method \(\overrightarrow{Q_iP_{j}^{c}}\times \overrightarrow{Q_iP_{j+1}^{c}}\le 0\).

During navigation, this vector method is also used to verify if the goal endpoint lies within the convex region. If so, this endpoint is directly used as the long-term reference point \(Q_{t}^{l}\) , which promotes early training by facilitating the acquisition of high-reward trajectories. Additionally, we introduce a final state stop cost \(\mathrm{cost}_{stop}\), into the MPC's cost function when the goal is within the convex region. This cost aims to minimize both velocity \(V_N=\left( v_{x}^{N},v_{y}^{N} \right)\) and acceleration \(A_N=\left( a_{x}^{N},a_{y}^{N} \right)\) at the destination, ensuring a controlled and safe termination of movement. 

The short-term and long-term reference points are used to guide the positions of the first and last control period states predicted by the MPC. The cost function incorporates discrepancies between MPC-predicted points and reference points to ensure fidelity to the reference trajectory. Ultimately, the MPC is defined as a quadratic programming problem, formulated in Equation \ref{eq:MPC_formulation}, optimizing the control sequence \(u_{0:N-1}^{*}\). \(w_{track}\) and \(w_{smooth}\) are the weights of the tracking error term and the smoothing term in the cost function respectively.  \(w_{vend}\) and \(w_{aend}\) are the final state velocity and acceleration weights of the optimised \(\mathrm{cost}_{stop}\). The optimization process generates a sequence of MPC-optimized points \(Q_{i}^{*}, i=\left\{ 1,...,N \right\}\), as depicted in Fig. \ref{fig:Observation Vector}.

\begin{equation}\label{eq:MPC_formulation}
\begin{aligned}
	\min_{u_{0:N-1}} &w_{track}\left( \left\| Q_1-Q_{t}^{s} \right\| ^2+\left\| Q_N-Q_{t}^{l} \right\| ^2 \right) \\
	&+w_{smooth}\sum_{k=0}^{N-1}{\left\| u_k \right\| ^2}+\cos t_{stop}\\
         \mathrm{cost}_{stop} &= \begin{cases}
            \begin{split}
                &w_{vend}\| V_N \|^2 \\
                &+ w_{aend}\| A_N \|^2
            \end{split} & \text{if goal in convex}\\
            0 & \text{otherwise}
        \end{cases}\\
	\mathrm{s}.\mathrm{t}.\quad &x_0=x_{init}\\
	&x_i=Fx_{i-1}+Gu_{i-1},\\
	&\overrightarrow{Q_iP_{j}^{c}}\times \overrightarrow{Q_iP_{j+1}^{c}}\le 0,\\
	&u_{i-1}\in \mathcal{U}, \quad \bar{x}_i\in \mathcal{S},\\
	&\forall i\in \{1,...,N\};\forall j=\left\{ 1,...,rnum_v-1 \right\}\\
\end{aligned}
\end{equation}

\subsection{Observation Space Formulation}\label{subsec:Observation Space}
\begin{figure}[htbp]
    \centering
    \includegraphics[width=8cm]{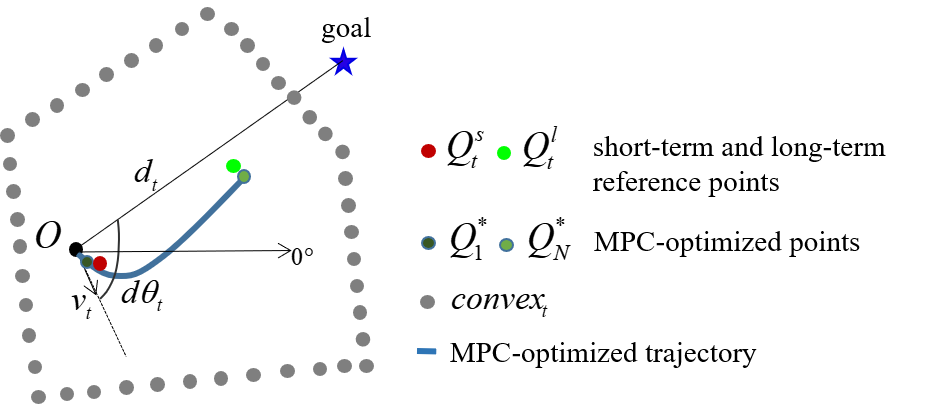}
    \caption{Schematic of the Observation Vector}
    \label{fig:Observation Vector}
\end{figure}

At time \(t\), the observation for the agent \(o_t\), as illustrated in Equation \ref{eq:Observation} and Figure \ref{fig:Observation Vector}, includes: the Euclidean distance \(d_t\) from robot to navigation goal, the velocity magnitude \(v_t\), the angular deviation \(d\theta_t\) between robot's velocity direction and the line to goal, the short-term and long-term reference points \(Q_{t-1}^{s}\) and \(Q_{t-1}^{l}\) from the previous frame's action, the MPC-optimized points \(Q_{1}^{*}\) and \(Q_{N}^{*}\) (defined in Section \ref{subsec:MPC}), and the convex region \(convex_t\).
\begin{equation}
o_t=\left( convex_t,d_t,v_t,d\theta _t,Q_{t-1}^{s},Q_{t-1}^{l},Q_{1}^{*},Q_{N}^{*} \right)
\label{eq:Observation}
\end{equation}

Due to the variability in shape of the convex obstacle-free region (\(convex_t\)) derived from different point cloud data, the number of vertices in convex polygons is not constant. This poses a challenge for neural networks, which require fixed dimension input. To address this issue, the number of vertices in the calculated convex region, denoted as \(num_v\) , is adjusted to match the predefined number \(rnum_v\): vertices are added through interpolation if the count \(num_v\) is less than \(rnum_v\), or reduced through sparsification if \(num_v\) exceeds \(rnum_v\).

The agent can extract motion information about dynamic obstacles and its own historical trajectories from continuous frames implicitly. The agent's state, denoted as \(s_t\), a concatenation of observations from three consecutive frames \(\left(o_{t-2}, o_{t-1}, o_t\right)\). The dimension of this state representation is set to \(6rnumv + 33\). This representation contains rich spatiotemporal information that is crucial for the agent's decision making process.

\subsection{Reward Function Formulation}\label{subsec:Reward Function}

The reward function plays a crucial role by setting the missions for agent. Agent’s policy is formulated based on the expectation of future rewards, making reward function a critical mechanism for conveying tasks to the agent.

A positive reward \(r_{success}\) is awarded to the agent when its distance to the goal \(d_t\) falls below \(d_{th}\).
\begin{equation}
r_{t}^{s}=\begin{cases}
	r_{success} & \text{if } d_t<d_{th}, \\
	0 & \text{otherwise}.
\end{cases}
\label{eq:r_s}
\end{equation}

The collision penalty \(r_{t}^{o}\), derived from distance data \(scan\) collected by LiDAR, is dynamically adjusted. The penalty decreases exponentially as the distance increases. This penalty is controlled by a positive decay factor \(w_{obs}\) and two negative terms \(r_{obs}\) and \(r_{collision}\).
\begin{equation}
r_{t}^{o}=\begin{cases}
	r_{collision} & \text{if } \min(\text{scan})\le 0\\
	r_{obs}e^{-w_{obs}\min(\text{scan})} & \text{if } \min(\text{scan})\le 2\\
	0 & \text{otherwise}
\end{cases}
\label{eq:r_o}
\end{equation}



To mitigate sparse reward, we introduce the reward \(r_{t}^{a}\) , based on the Euclidean distance \(d_t\) between the robot and its destination which controlled by a positive factor \(w_{approach}\). 
\begin{equation}
r_{t}^{a}=\begin{cases}
	0 & \text{if } t=1\\
	w_{approach}(d_t-d_{t-1}) & \text{otherwise}
\end{cases}
\end{equation}
We also tried adding guide reward based on global path to \(r_{t}^{a}\).

Short-term and long-term reference points, \(Q_{t}^{s}\) and \(Q_{t}^{l}\), discussed in Section~\ref{subsec:Action Space}, might not be practical to reach. However, the MPC-optimized positions \(Q_{i}^{*}\) , outlined in Section~\ref{subsec:MPC}, comply with kinematic constraints and provide more feasible references. Thus, to penalize the discrepancy between reference points and MPC-optimized points, we introduce \(r_{t}^{f}\).This reward encourages the generation of reachable reference points.This penalty is controlled by two negative penalty factor \(w_{feasible}^{s}\) and \(w_{feasible}^{l}\).
\begin{equation}
r_{t}^{f}=w_{feasible}^{s}\left\| Q_{t}^{s}-Q_{1}^{*} \right\| ^2+w_{feasible}^{l}\left\| Q_{t}^{l}-Q_{N}^{*} \right\| ^2
\end{equation}

Significant changes in reference points across frames can result in deviations between MPC-optimized trajectories at consecutive steps. Consequently, the agent requires increased control efforts to mitigate the influence of previous frames. To address this, the penalty \(r_{t}^{c}\) is introduced, which can also facilitate rapid learning of smooth direction changes during the initial training phase.This penalty is controlled by two negative penalty factor \(w_{change}^{s}\) and \(w_{change}^{l}\).
\begin{equation}
r_{t}^{c}=\begin{cases}
	0 & \text{if } t=1\\
	\begin{aligned}
		&w_{change}^{s}\| Q_{t}^{s}-Q_{t-1}^{s} \|^2 \\
		&+ w_{change}^{l}\| Q_{t}^{l}-Q_{t-1}^{l} \|^2
	\end{aligned} & \text{otherwise}
\end{cases}
\end{equation}

Ultimately, the fixed per-step penalty, denoted as \(r_t^e\), is introduced.

The final reward function, denoted as \(r_t\), is composed of the above six components. These components can be adjusted or selectively ignored during phases of training.
\begin{equation}
r_t=r_{t}^{e}+r_{t}^{s}+r_{t}^{a}+r_{t}^{o}+r_{t}^{c}+r_{t}^{f}
\end{equation}

\subsection{Network Architecture}
\begin{figure}[htbp]
    \centering
    \includegraphics[width=8cm]{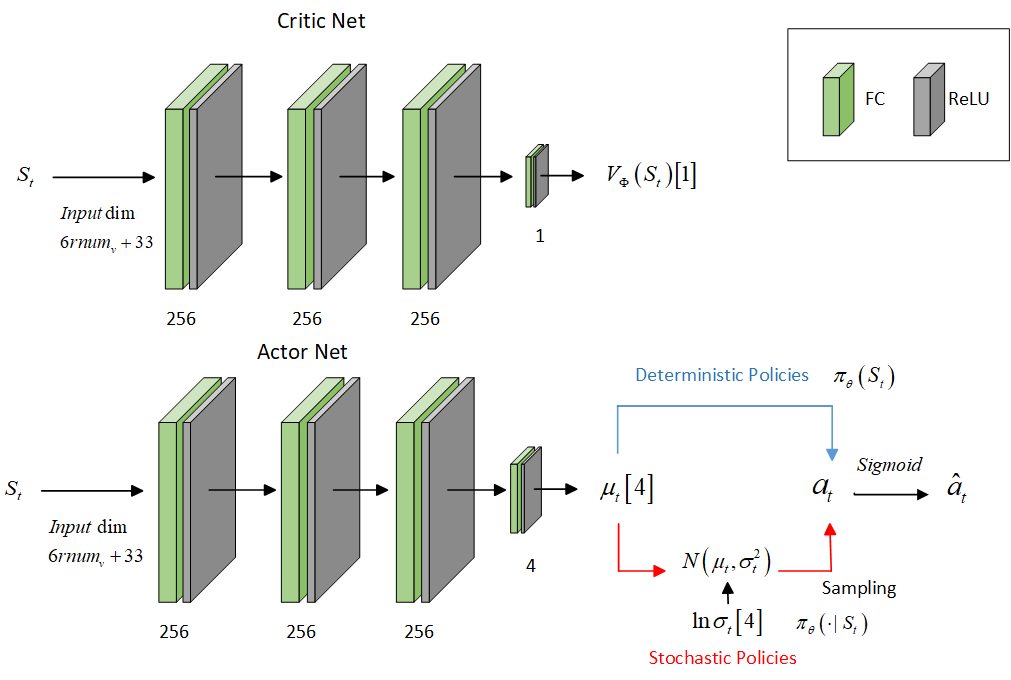}
    \caption{Network Architecture: The value network \(V_{\phi}\) and policy network \(\pi_{\theta}\) are structured as four-layer fully connected networks. The policy network is augmented with a logarithmic standard deviation parameter (\(\ln \sigma_t\)) for each action dimension. This configuration primarily facilitates the generation of stochastic policies by allowing actions to be sampled from a Gaussian distribution.}
    \label{fig:Network Architecture}
\end{figure}

The convex obstacle-free region efficiently removes redundant information from the raw LiDAR point cloud, including noise, duplicate points, and distant obstacles. This preprocessing step allows for the implementation of a more streamlined network structure for fitting policy and value functions, as depicted in Figure \ref{fig:Network Architecture}. Given that separate networks can yield better performance in practice \cite{decoupling_network}, the policy and value network are updated independently, without sharing parameters.

\section{EXPERIMENTS}
In this section, we introduce our training process along with the definition of seven Stages that differ in complexity (Table \ref{table:obstacle_parameters}). These stages are specifically designed for curriculum learning and subsequent evaluation. Following this, we evaluate our method in three parts:

(1) To demonstrate the advantages of our method in terms of action and state space design, we conduct an ablation study. This study compares our method with four different DRL navigation methods constructed with various action and state spaces (Section \ref{subsec:ablation_action}).

(2) To validate the effectiveness of our reward function design, we conduct another ablation study. This study conducts a comparison between our method and three other DRL navigation methods with different reward configurations (Section \ref{subsec:ablation_reward}).

(3) In order to further analyze our method's advantages over non-DRL navigation methods in dynamic and crowded environments, we conduct a comparison between our method and the TEB \cite{teb} in test scenarios (Section \ref{subsec:evaluation}).

\subsection{Training Procedure}
Our method employs a multi-stage training strategy that progresses from simple to complex scenarios, following the principle of curriculum learning. This strategy aims to enhance the agent's learning efficiency by optimizing the sequence of acquiring experience. The agent first learns basic navigation movements in an obstacle-free environment (Stage 1) initially and gradually being exposed to increasingly complex static and dynamic obstacles.

\subsubsection{Stage Definition}
The parameters of static and dynamic obstacles over seven Stages for staged training and subsequent evaluation are presented in Table \ref{table:obstacle_parameters}. Static obstacles are randomly generated polygons with three or four sides, where the maximum area of these polygons does not exceed 2 square meters. The dynamic obstacles are circular in shape.

In order to assess the performance of the trained agent, we generate 1,000 test scenarios for each Stage, from 2 to 7. These scenarios are generated by utilizing random seeds that are not included in the training dataset to configure the simulator.

\begin{table}[htbp]
\caption{The Parameters of Static and Dynamic Obstacles in Seven Stages for Staged Training and Evaluation}
\label{table:obstacle_parameters}
\begin{center}
\begin{tabular}{|c|c|c|c|c|c|}
\hline
Stage & \makecell{Size\\(m)} & \makecell{Static\\Obstacles\\Count} & \makecell{Dynamic\\Obstacles\\Count} & \makecell{Dynamic\\Obstacle\\Radius\\Range\\(m)} & \makecell{Dynamic\\Obstacle\\Speed\\Range\\(m/s)} \\
\hline
1 & 20$\times$30 & 0 & 0 & - & - \\
\hline
2 & 20$\times$30 & 10 & 0 & - & - \\
\hline
3 & 20$\times$30 & 10 & 5 & 0.2-0.3 & 0.3 \\
\hline
4 & 20$\times$30 & 10 & 10 & 0.2-0.3 & 0.3 \\
\hline
5 & 10$\times$10 & 0 & 10 & 0.1-0.4 & 0.3-0.6 \\
\hline
6 & 10$\times$10 & 0 & 20 & 0.1-0.4 & 0.3-0.6 \\
\hline
7 & 10$\times$10 & 0 & 30 & 0.1-0.4 & 0.3-0.6 \\
\hline
\end{tabular}
\end{center}
\end{table}

\subsubsection{Implementation Details}
Our study utilizes the Arena-Rosnav framework \cite{Arena-Rosnav}, developed within the ROS architecture, to enable reliable robotic navigation solutions. We also adopt the implementation of Proximal Policy Optimization (PPO) \cite{PPO} from Elegant-RL \cite{elegantrl}, which provides tools for network customization and training process monitoring. The simulation experiments are conducted on a high-performance PC equipped with an Intel Core i7-770to0K CPU, an Nvidia RTX 2080Ti GPU, and 32GB of RAM.In addition, we tried multi-GPU training experiments on a computer with an i7-6800k CPU and two Nvidia RTX 1080 Ti GPUs.

\subsection{Ablation Study on State and Action Space}\label{subsec:ablation_action}
To verify the effectiveness of the proposed method in terms of action and state space design, we conducted comparison experiments against four methods:

Design 1: Utilizing the conventional end-to-end architecture where the observation \(o_t\) includes LiDAR data \(scan_{t}^{'}\) . Action is defined as a set of two-dimensional velocities \(\left( v_x,v_y \right)\) . The relationship between the output of the policy network \(\left( \alpha _t,\beta _t \right)\) and \(\left( v_x,v_y \right)\) is detailed in (\ref{eq:Design 1}).
\begin{equation}
\begin{aligned}
o_t &= \left( scan_{t}^{'},d_t,v_t,d\theta _t \right) \\
\hat{a}_t &= \tanh \left( a_t \right) =\left\{ \left( \alpha _t,\beta _t \right) |\alpha _t,\beta _t\in \left( -1,1 \right) \right\} \\
v_x &= v_{min}+\left( v_{max}-v_{min} \right) \cdot \alpha _t \\
v_y &= v_{min}+\left( v_{max}-v_{min} \right) \cdot \beta _t
\end{aligned}
\label{eq:Design 1}
\end{equation}

Design 2: The state observation \(o_t\) is updated to (\ref{eq:Design 2}), the convex obstacle-free region is taken into account. The action, consisting of \((v_x,v_y)\), consistent with Design 1.
\begin{equation}
o_t=\left( convex_t,d_t,v_t,d\theta _t \right)
\label{eq:Design 2}
\end{equation}

Design 3: The action is simplified to \((Q_{t}^{l})\), focusing exclusively on the long-term reference point. The calculation process of a single \((Q_{t}^{l})\) is similar to the short-term reference point \((Q_{t}^{s})\) in Section \ref{subsec:Action Space}, except that single \((Q_{t}^{l})\) is sampled from the long-term action space. In addition, Design 3 simplifies the state observation in (\ref{eq:Design 3}) by removing terms related to \(Q_{t}^{s}\).
\begin{equation}
o_t=\left( convex_t,d_t,v_t,d\theta _t,Q_{t-1}^{l},Q_{N}^{*} \right)
\label{eq:Design 3}
\end{equation}

Design 4: The observation \(o_t\) includes raw LiDAR data \(scan_{t}^{'}\), as shown in (\ref{eq:Design 4}). The action is still \(\left(Q_{t}^{l},Q_{t}^{s}\right)\), but calculated within the intersection of the point cloud's coverage and the robot's kinematic limits. Given that \(scan_{t}^{'}\) is non-convex and cannot be used as a constraint for convex optimization, the constraint that the optimized trajectory must be within the convex region is removed during the MPC calculation.
\begin{equation}
o_t=\left( scan_{t}^{'},d_t,v_t,d\theta _t,Q_{t-1}^{s},Q_{t-1}^{l},Q_{1}^{*},Q_{N}^{*} \right)
\label{eq:Design 4}
\end{equation}

After training from Stage 2 to 4, the performance of our method and Designs 1 to 4 in test scenarios is shown in Table \ref{table:table_ablation_study}. 

\begin{table}[htbp]
\caption{Ablation Study on Action and State Space Designs: Comparative results of our method and Design 1 to 4 Across 1,000 test scenarios for each of the Stages from 2 to 4}
\label{table:table_ablation_study}
\centering
\begin{tabular}{|c|c|c|c|c|c|}
\hline
Stage & Method & \makecell{Success\\Rate\\(\%)} & \makecell{Time\\(s)} & \makecell{Distance\\(m)} & \makecell{Speed\\(m/s)} \\
\hline
  & Design 1 & 76 & 4.0 & 11.28 & 2.87 \\
  & Design 2 & 76 & 4.0 & 11.11 & 2.8 \\
2 & Design 3 & \textbf{90.3} & 9.0 & 15.26 & 1.72 \\
  & Design 4 & 83 & 5.0 & 11.66 & 2.23 \\
  & Ours     & 89.2 & 5.5 & 12.19 & 2.18 \\
\hline
& Design 1 & 79.9 & 4.0 & 11.37 & 2.9 \\
& Design 2 & 79 & 4.0 & 11.4 & 2.88 \\
3 & Design 3 & 87.8 & 8.8 & 15.42 & 1.77 \\
& Design 4 & 83.5 & 4.9 & 11.59 & 2.24 \\
& Ours & \textbf{89.1} & 5.9 & 13.04 & 2.17 \\
\hline
& Design 1 & 80.9 & 3.9 & 11.25 & 2.95 \\
& Design 2 & 75.1 & 3.8 & 10.78 & 2.89 \\
4 & Design 3 & 84.5 & 9.0 & 15.13 & 1.69 \\
& Design 4 & 83.7 & 4.8 & 11.19 & 2.2 \\
& Ours & \textbf{85.5} & 5.8 & 12.51 & 2.1 \\
\hline
\end{tabular}
\end{table}

In terms of action space design, Design 3 simplifies action to a single long-term reference point. Although Design 3 has an average navigation success rate of 87.53\%, close to our method's 87.93\%, Design 3's average navigation time is 3.2s longer and the average navigation distance is 2.69m longer than our method in the three Stages. On the other hand, compared with Design 1 and 2, the navigation success rate of our method is 9\% and 11.23\% higher respectively.

In terms of state space design, compared with Design 1 and Design 4 that directly use raw LiDAR data as the observation, our method's average navigation success rate is 9\% and 4.53\% higher, respectively. The ablation study demonstrates the effectiveness of our method in the design of the action and state space.


\subsection{Ablation Study on Reward Function}\label{subsec:ablation_reward}
To assess the effectiveness of our reward function design, particularly the \(r_{t}^{c}\) and \(r_{t}^{f}\) items, as detailed in Section~\ref{subsec:Reward Function}, we conducted the ablation study with four different reward configurations and evaluated them across 1000 test scenarios from Stage 5 to 7. The results are presented in Table \ref{table:reward_function_ablation}.

The configuration of reward function \(r_{t1}\) used in our method includes all items.
\begin{equation}
r_{t1}=r_{t}^{e}+r_{t}^{s}+r_{t}^{a}+r_{t}^{o}+r_{t}^{c}+r_{t}^{f}
\label{eq:r_{t1}}
\end{equation}

Reward function \(r_{t2}\) is based on \(r_{t1}\) but excludes the
reference points change penalty (\(r_{t}^{c}\)).
\begin{equation}
r_{t2}=r_{t}^{e}+r_{t}^{s}+r_{t}^{a}+r_{t}^{o}+r_{t}^{f}
\label{eq:r_{t2}}
\end{equation}

Reward function \(r_{t3}\) is based on \(r_{t1}\) but excludes the
MPC-optimized points error penalty (\(r_{t}^{f}\)).
\begin{equation}
r_{t3}=r_{t}^{e}+r_{t}^{s}+r_{t}^{a}+r_{t}^{o}+r_{t}^{c}
\label{eq:r_{t3}}
\end{equation}

Reward function \(r_{t4}\) excludes both the \(r_{t}^{c}\) and the
\(r_{t}^{f}\) from \(r_{t1}\).
\begin{equation}
r_{t4}=r_{t}^{e}+r_{t}^{s}+r_{t}^{a}+r_{t}^{o}
\label{eq:r_{t4}}
\end{equation}

\begin{table}[htbp]
\caption{Ablation Study on Reward Function Designs: Comparative results of four reward functions across 1,000 test scenarios for each of the Stage from 5 to 7}
\label{table:reward_function_ablation}
\centering
\begin{tabular}{|c|c|c|c|c|c|c|}
\hline
\makecell{Stage} & \makecell{Reward} & \makecell{Success\\Rate\\(\%)} & \makecell{Time\\(s)} & \makecell{Distance\\(m)} & \makecell{Speed\\(m/s)} & \makecell{Total\\Abs\\Acc} \\
\hline
\multirow{4}{*}{5} & $r_{t1}$        & 87.1       & 2.9      & 4.59     & 1.57  & 26475.46      \\
                   & $r_{t2}$        & \textbf{87.8}       & 2.9      & 4.65     & 1.57  & 27600.56      \\
                   & $r_{t3}$        & 84.8       & 3.4      & 4.38     & 1.40  & 27781.34      \\
                   & $r_{t4}$        & 86.7       & 2.9      & 4.65     & 1.58  & \textbf{26460.66}      \\
\hline
\multirow{4}{*}{6} & $r_{t1}$        & \textbf{78.1}       & 2.9      & 4.12     & 1.42  & \textbf{19507.60}      \\
                   & $r_{t2}$        & 77.3       & 2.8      & 4.10     & 1.43  & 20397.66      \\
                   & $r_{t3}$        & 73.1       & 3.4      & 3.85     & 1.25  & 21988.81      \\
                   & $r_{t4}$        & 74.6       & 3.0      & 4.35     & 1.46  & 19950.77      \\
\hline
\multirow{4}{*}{7} & $r_{t1}$        & 68.5       & 3.1      & 4.22     & 1.36  & 6348.57       \\
                   & $r_{t2}$        & \textbf{69.9}       & 3.0      & 4.15     & 1.37  & 7218.53       \\
                   & $r_{t3}$        & 67.1       & 3.6      & 3.98     & 1.19  & 6080.69       \\
                   & $r_{t4}$        & 65.3       & 2.9      & 3.88     & 1.36  & \textbf{5668.30}       \\
\hline
\end{tabular}
\end{table}

When evaluating the impact of composite reward functions on motion smoothness, Total Absolute Acceleration (Total Abs Acc) is a crucial metric. Total Abs Acc is the sum of the agent's absolute acceleration taken across steps within successful episodes. 

The ablation study clarifies the importance of the MPC-optimized points error penalty (\(r_{t}^{f}\)) for high success rates. This is evident in the performance of \(r_{t1}\)'s performance, which achieved success rates of 87.1\%, 78.1\%, and 68.5\% across stages. When \(r_{t}^{f}\) is excluded in \(r_{t3}\), the success rates decline to 84.8\%, 73.1\%, 67.1\%. Conversely, the reference points change penalty (\(r_{t}^{c}\)) slightly affects success rates, but it is significant for motion smoothness. The \(r_{t2}\) shows increased Total Abs Acc (27600.56 in Stage 5) compared to \(r_{t1}\)(26475.46 in Stage 5), indicating smoother navigation with \(r_{t}^{c}\) included. 

\subsection{Evaluation}\label{subsec:evaluation}
Following Stages 5-7's training in tight, dynamic crowded environments, both our method and TEB method are evaluated in 1,000 test scenarios for each Stage, with results detailed in Table \ref{table:table_comparison_TEB}.In order to use the same kinematic model as our method, we debug and experiment on the basis of omni-directional TEB\footnote{\url{https://github.com/pingplug/teb_local_planner/tree/omni_type}}.

\begin{table}[htbp]
\caption{Comparative Results of Our Method and Teb Method Across 1,000 Test Scenarios for Each of the Stage from 5 to 7}
\label{table:table_comparison_TEB}
\centering
\begin{tabular}{|c|c|c|c|c|c|}
\hline
Stage & Method & \makecell{Success\\Rate\\(\%)} & \makecell{Time\\(s)} & \makecell{Distance\\(m)} & \makecell{Speed\\(m/s)} \\
\hline
\multirow{2}{*}{5}     & TEB\cite{teb}    & 77.1       & 4.4      & 4.64         & 1.11        \\
                       & Ours   & \textbf{87.1}       & 2.9      & 4.59        & 1.57        \\
\hline
\multirow{2}{*}{6}     & TEB    & 65.2       & 4.0      & 4.21         & 1.10        \\
                       & Ours   & \textbf{78.1}       & 2.9      & 4.12         & 1.42        \\
\hline
\multirow{2}{*}{7}     & TEB    & 57.8       & 3.8      & 3.87         & 1.07        \\
                       & Ours   & \textbf{68.5}       & 3.1      & 4.22         & 1.36        \\
\hline
\end{tabular}
\end{table}

Our method demonstrates higher navigation success rates than the TEB method across all three Stages, outperforming TEB by 10.0\%, 12.9\%, and 10.7\%, respectively. The results highlight the ability of our method to avoid obstacles in highly dynamic environments.

\begin{figure}[htbp]
    \centering
    \includegraphics[width=6cm]{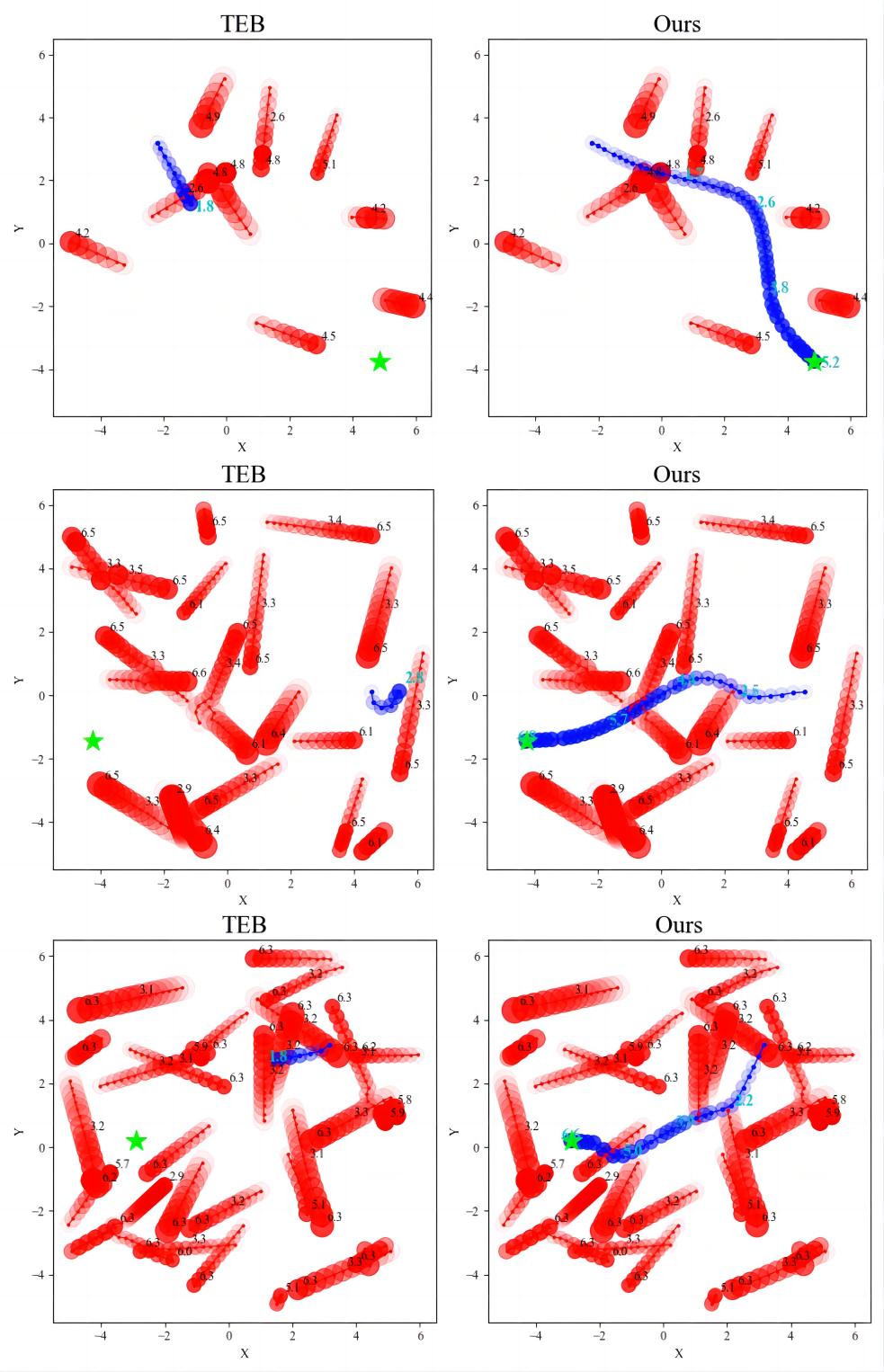}
    \caption{Comparative Navigation Process: Our Method and TEB in Selected Test Scenarios from Stages 5, 6, 7. The three images, arranged from top to bottom, respectively depict the robot's navigation process during the test scenarios of Stages 5, 6, and 7. Dark blue line marking robot trajectory, a green star for the goal, and cyan numbers indicate the time used by robot to move to this position, in seconds. Red lines with adjacent black numbers represent dynamic obstacles' paths and timing move to this position, respectively.}
    \label{fig:Comparative TEB}
\end{figure}

In order to compare the navigation behavior of our method with TEB, several scenarios from each stage are visualized in Figure \ref{fig:Comparative TEB}. While TEB struggles with predicting the movement of dynamic obstacles, leading to potential collisions. In contrast, our method moves through available gaps, allowing for more efficient obstacle avoidance.

\subsection{Qualitative Analysis}\label{subsec:qualitative}

\begin{figure}[htbp]
    \centering
    \includegraphics[width=5cm]{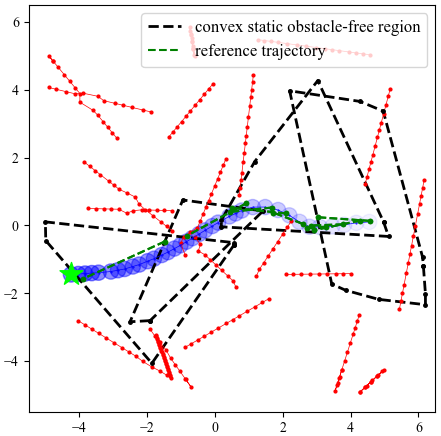}
    \caption{Navigation Process of Our Method in a Test Scenario at Stage 6 for Qualitative Analysis: Omitting the radius and timing of dynamic obstacles, as well as the robot’s timing data. The convex obstacle-free regions are indicated by black dashed lines. And a portion of reference trajectories are indicated by green dashed lines. The reference trajectory is formed by connecting long-term and short-term reference points, as detailed in Section \ref{subsec:Action Space}.}
    \label{fig:Qualitative Analysis}
\end{figure}
Figure \ref{fig:Qualitative Analysis} illustrates the navigation process, reflecting the iterative trajectory optimization procedure depicted in Figure \ref{fig:Iterative Trajectory Optimization}. The robot updates the convex obstacle-free region with the latest LiDAR data. Subsequently, long-term and short-term reference points are sampled within this convex region to form the reference trajectory. The MPC then tracks this reference trajectory to solve the local navigation problem, until the goal endpoint is within the latest convex region and is directly adopted as the long-term reference point. The process aims to promote safety by constraining both the reference trajectory and robot motion within the overlapping convex regions.
\section{CONCLUSIONS}

This paper introduces a DRL-based navigation framework, which consists of the design of action and state spaces, reward function and network architecture. In terms of action space design, our method integrates lidar-generated convex obstacle-free region to formulate the action space that includes both short-term and long-term reference points. For the reward function, we devise a composite reward function enriched with intermediate rewards. The experimental results conclusively demonstrate that the proposed method significantly enhances robotic performance in both static and dynamic environments, notably excelling in dynamic and crowded environments.

\addtolength{\textheight}{-12cm}   







\bibliographystyle{IEEEtran}
\bibliography{references}

\end{document}